\pdfoutput=1

\documentclass[11pt]{article}

\usepackage[]{acl}

\usepackage{times}
\usepackage{latexsym}

\usepackage[T1]{fontenc}
\usepackage{amsfonts}

\usepackage[utf8]{inputenc}

\usepackage{microtype}

\usepackage{amsmath}
\usepackage{graphicx}

\usepackage{float}
\usepackage{pgfplots}
\usepackage{subfigure}
\usepackage{booktabs}
\usepackage{multicol}
\usepackage{multirow}
\usepackage{arydshln} 
\usepackage{CJKutf8}
\usepackage{hyperref}
\usepackage{bbding}

\title{Contextualized Semantic Distance between Highly Overlapped Texts}

\author{Letian Peng$^{1}$, Zuchao Li$^{2,*}$, and Hai Zhao$^{3}$\thanks{$\ $  Corresponding author. This work was supported by Key Projects of the National Natural Science Foundation of China (U1836222 and
61733011), the Fundamental Research Funds for the Central Universities and the Fundamental Research Funds for the Central Universities (No. 2042023kf0133).}\\
$^{1}$Department of Computer Science and Engineering, University of California San Diego \\
$^{2}$National Engineering Research Center for Multimedia Software, \\
School of Computer Science, Wuhan University, Wuhan, 430072, P. R. China \\
$^{3}$Department of Computer Science and Engineering, Shanghai Jiao Tong University\\
  {\tt \small lepeng@ucsd.edu, zcli-charlie@whu.edu.cn, zhaohai@cs.sjtu.edu.cn}}
  
\begin{document}
\maketitle
\begin{abstract}

Overlapping frequently occurs in paired texts in natural language processing tasks like text editing and semantic similarity evaluation. Better evaluation of the semantic distance between the overlapped sentences benefits the language system's understanding and guides the generation. Since conventional semantic metrics are based on word representations, they are vulnerable to the disturbance of overlapped components with similar representations. This paper aims to address the issue with a mask-and-predict strategy. We take the words in the longest common sequence (LCS) as neighboring words and use masked language modeling (MLM) from pre-trained language models (PLMs) to predict the distributions in their positions. Our metric, Neighboring Distribution Divergence (NDD), represents the semantic distance by calculating the divergence between distributions in the overlapped parts. Experiments on Semantic Textual Similarity show NDD to be more sensitive to various semantic differences, especially on highly overlapped paired texts. Based on the discovery, we further implement an unsupervised and training-free method for text compression, leading to a significant improvement on the previous perplexity-based method. The high compression rate controlling ability of our method even enables NDD to outperform the supervised state-of-the-art in domain adaption by a huge margin. Further experiments on syntax and semantics analyses verify the awareness of internal sentence structures, indicating the high potential of NDD for further studies.\footnote{Our code is released at \href{https://github.com/Stareru/NeighboringDistributionDivergence/}{github.com/Stareru/\\NeighboringDistributionDivergence/}}


\end{abstract}

\section{Introduction}

 Comparison between highly overlapped sentences exists in many natural language processing (NLP) tasks, like text rewriting \citep{DBLP:conf/emnlp/LiuCLZZ20} and semantic textual similarity \cite{DBLP:conf/naacl/ZhelezniakSSH19}. A reliable evaluation of these paired sentences will benefit controllable generation and precise semantic difference understanding. 

Conventional metrics, like the cosine similarity (S$_C$), have been popular for semantics similarity evaluation. Nevertheless, we find the evaluating capability of S$_C$ severely degrades when the overlapping ratio rises. \citeauthor{DBLP:journals/corr/abs-1909-03223} try to introduce the difference between perplexity ($\Delta$PPL) to describe the semantic distance. Unfortunately, $\Delta$PPL suffers from the word frequency imbalance. Also, many sentences share a similar PPL. 

\begin{figure}
    \centering
    \includegraphics[width=0.5\textwidth]{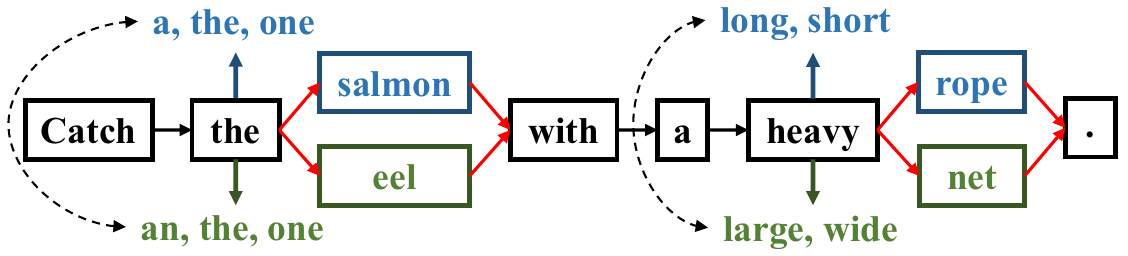}
    \caption{The comparison between two possible text scenarios with shared components. Mask-and-predicting neighboring words attenuates disturbance from overlapping when evaluating semantic distance.}
    \label{fig:ex}
\end{figure}

\begin{figure*}
    \centering
    \includegraphics[width=0.99\textwidth]{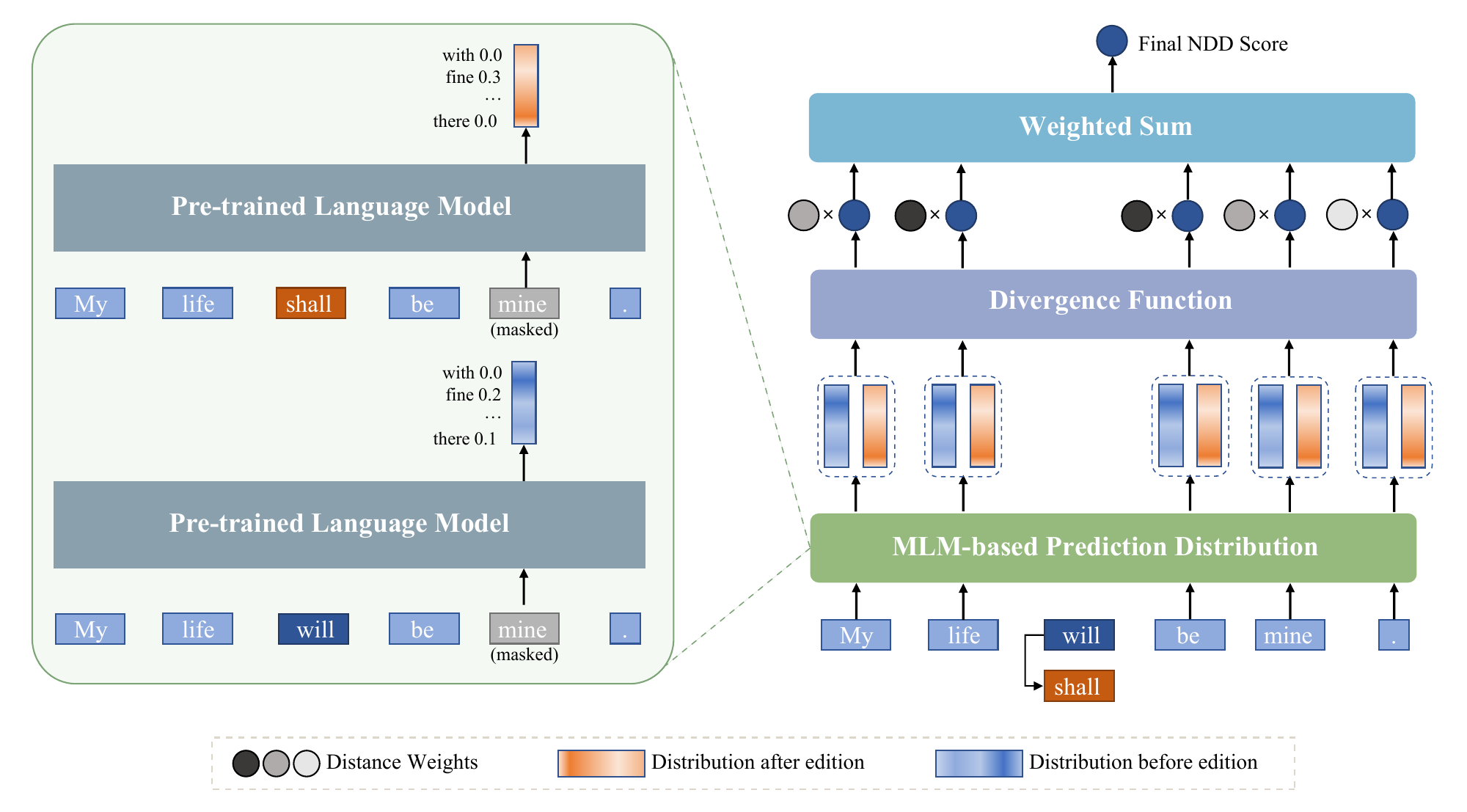}
    \caption{Calculating procedure for Neighboring Distribution Divergence.}
    \label{fig:ndd}
\end{figure*}

Based on the failure of S$_C$, we hypothesize that the evaluation is disturbed by the overlapped components, which share similar representations in the paired sentences. We thus intend to mitigate the disturbance and thus propose a mask-and-predict strategy to attenuate the disturbance from overlapped words. Compared to directly using the word representations for comparison, we discover that using predicted distributions from masked language modeling (MLM) results in better evaluation. Taking Figure~\ref{fig:ex} as the instance, unmasked comparison involves similar representations of \textit{the} and \textit{heavy} in both sentences since the encoder can see these words and encode with their information. But when these words are masked, the MLM has to predict the distributions considering the contextual difference. While using representations results in a trivial \textit{heavy}-\textit{heavy} comparison, the difference of distributions (between candidates \textit{long}, \textit{short} and \textit{large},\textit{wide}) better indicates how the contextual semantics changes.

Thus, we are motivated to propose a new metric, Neighboring Distribution Divergence, which compares predicted MLM distributions from pre-trained language models (PLMs) and uses the the divergence between them to represent the semantic distance. We take the overlapped words in the longest common sequence (LCS) between the paired sentences as neighboring words for the divergence calculation. We conduct experiments on semantic textual similarity and text compression. Experiment results verify NDD to be more sensitive to precise semantic differences than conventional metrics like S$_C$. Experiments on the Google dataset show our method outperforms the previous PPL-based baseline by around $10.0$ on F1 and ROUGE scores. Moreover, the NDD-based method enjoys outstanding compression rate controlling ability, which enables it to outperform the supervised state-of-the-art by $18.8$ F1 scores when adapting to Broad News Compression Corpus in a new domain. The cross-language generality of NDD is also verified by experiments on a Chinese colloquial Sentence Compression dataset. 

We further use syntax and semantics analyses to test NDD's awareness of the sentence's internal structure. Our experiments show that NDD can be applied for accurate syntactic subtree pruning and semantic predicate detection. Results from our analyses verify the potential of NDD on more syntax or semantics-related tasks. Our contributions are summarized as follows:

\begin{itemize}
    \item We address the component overlapping issue in text comparison by using a mask-and-predict strategy and proposing a new metric, Neighboring Distribution Divergence. 
    \item We use semantic tests to verify NDD to be more sensitive to various semantic differences than previous metrics. 
    \item NDD-based training-free algorithm has strong performance and compression rate controlling ability. The algorithm sets the new unsupervised state-of-the-art on the Google dataset and outperforms the supervised state-of-the-art by a sharp margin on the Broad News Compression dataset. 
    \item Further syntax and semantics analyses show NDD's awareness of internal structures in sentences.
\end{itemize}

\section{Neighboring Distribution Divergence}

\subsection{Background}

Before the main discussion, we first recall the definition of perplexity and cosine similarity as the basis for further discussion. 

\paragraph{Perplexity} For a sentence with $n$ words (more specifically, subwords) $W = [w_1, w_2, \cdots, w_n]$, perplexity refers to the average of log possibility for each word to exist in $W$. If the perplexity is evaluated by an MLM-based PLM, then the existing possibility is represented by the predicting distribution on the masked position.

\begin{equation*}
    \centering
    \begin{aligned}
    W_m &= [w_1, \cdots, w_{i-1}, \textrm{[MASK]}, w_{i+1}, \cdots, w_n]\\
    Q &= \textrm{PLM}^{MLM}(W_m), q_i = \textrm{softmax}(Q_i) \in \mathbb{R}^{c}\\
    p_i &= q_{i,\textrm{Idx}(w_i)}, \textrm{PPL} = \frac{1}{n}\sum_{i=0}^{n}-\log{(p_i)}
    \end{aligned}
\end{equation*}

The PLM predicts the distribution $Q$ for the masked word on $i$-th position. Then, the softmax function is used to get the probability distribution $Q$ where $q_j$ refers to the appearance possibility of $j$-th word in the $c$-word dictionary on $i$-th position. Here $\textrm{Idx}(\cdot)$ returns the index of word in the dictionary. The distribution predicting process is summarized as a function $\textrm{MLM}(\cdot)$ where $\textrm{MLM}(W, i) = q_i$.

As implausible words or structures will result in high perplexity, this metric can reflect some semantic information. Perplexity is commonly used to evaluate the plausibility of text and detect semantic errors in sentences. 

\paragraph{Cosine Similarity} For the a sentence pair $W_{x}$, $W_{y}$, a pre-trained encoder (like PLM or word embedding) encodes their contextual representations as $R_{x}$, $R_{y}$. We use PLM-based S$_C$ for experiments and follow the best-representing scenario in \cite{DBLP:conf/emnlp/GaoYC21} to use the CLS token as the sentence representation.

\begin{equation*}
    \centering
    \begin{aligned}
    &R_{x} = \textrm{PLM}(W_{x}), R_{y} = \textrm{PLM}(W_{y})\\
    &\textrm{S}_{C}(W_{x}, W_{y}) = \frac{R_{x}^{CLS} \cdot R_{y}^{CLS}}{||R_{x}^{CLS}||\times ||R_{y}^{CLS}||}
    \end{aligned}
\end{equation*}

\subsection{The Calculation Method}

This section will detail the steps involved in determining the Neighboring Distribution Divergence. Breaking down the term NDD, \textbf{Neighboring} refers to words contained within the longest common subsequence, \textbf{Distribution} is in reference to the predicted results of the Masked Language Model on those neighboring words, while \textbf{Divergence} signifies the disparity between the predicted distributions within the LCS of the pair of sentences under scrutiny.

We'll start with a sentence pair, denoted as $(W, W')$. The first step is to identify the LCS between these sentences, denoted as $W_{LCS}$. Words within this LCS will serve as our neighboring words for comparison. The Pretraining Language Model (PLM) is applied to each word in $W_{LCS}$ to predict their respective distributions using the MLM. Subsequently, a divergence function is employed to assess the distribution divergence between $W$ and $W'$ based on the same shared word. The divergence scores obtained are then assigned weights and totaled to produce the final NDD output.

The process can be mathematically expressed as:

\begin{equation*}
    \centering
    \begin{aligned}
    q_i &= \textrm{MLM}(W, i), q'_i = \textrm{MLM}(W', i)\\
    NDD &= \sum_{w \in W_{LCS}} a_w\text{F}_{div}(q_{\textrm{Idx}^d(w)}, q'_{\textrm{Idx}'^d(w)})
    \end{aligned}
\end{equation*}

In this equation, $\textrm{F}_{div}(\cdot)$ symbolizes a divergence function that calculates the divergence between distributions. The functions Idx$^d(\cdot)$ and Idx$'^d(\cdot)$ are used to identify the index of $w$ in sentences $W$ and $W'$ respectively. The term $a_w$ denotes the weight assigned to each word $w$, which inversely corresponds to its proximity to the nearest word outside the LCS.

\section{Semantic Distance Evaluation}

We conduct experiments on the test dataset of Semantic Textual Similarity Benchmark~\footnote{\href{http://ixa2.si.ehu.eus/stswiki/index.php/STSbenchmark}{http://ixa2.si.ehu.eus/stswiki/index.php/STSbenchmark}} (STS-B) to analyze the metrics. Multiple sentence pair similarity evaluation tasks are designed to compare the metric performance and investigate the metric property.

\begin{itemize}
    \item \textbf{Synonym-Antonym test} creates sentence pairs by replacing words with their synonyms and antonyms. Replacing by the synonym (antonym) results in a positive (negative) pair. 
    \item \textbf{Part-of-speech (POS) test} replaces words with ones that have the same (positive) or different (negative) parts-of-speech\footnote{Linguistic features in this paper are gotten by models from SpaCy. \href{https://spacy.io/}{https://spacy.io/}}. 
    \item \textbf{Term test} replaces verbs with ones in the same (positive) and different (negative) terms. 
    \item \textbf{Lemma test} replaces words with ones that have the same (positive) or different (negative) lemma root. 
    \item \textbf{Supervised test} uses the human-annotated scores for STS-B sentence pairs.
\end{itemize}

\begin{table}
    \centering
    \small
    \scalebox{1.0}{ 
    \begin{tabular}{lcccccc}
    \toprule
    Metric & Syn-Ant & POS & Term & Lemma & Sup. \\
    \midrule
    $\Delta$PPL & 5.3 & 2.8 & 8.7 & 7.9 & 11.2 \\
    S$_C$ & 7.8 & 8.1 & 9.1 & 0.0 & 20.8 \\
    NDD & \textbf{19.1} & \textbf{22.7} & \textbf{11.2} & \textbf{17.8} & 24.0 \\
    NDD + S$_C$ & 12.5 & 14.5 & 10.8 & 6.8 & \textbf{28.2} \\
    \bottomrule
    \end{tabular}
    }
    \caption{Text similarity evaluation on STS-B subset. We use Pearson Correlation as the evaluating metric. \textbf{Syn-Ant:} synonym-antonym test. \textbf{POS:} part-of-speech test. \textbf{Term:} verb term test. \textbf{Lemma:} lemma test. \textbf{Sup.:} Supervised STS test.}
    \label{tab:sts}
\end{table}

\definecolor{lust}{rgb}{0.9, 0.13, 0.13}
\definecolor{malachite}{rgb}{0.04, 0.85, 0.32}

We replace $20\%$ words for synonym-antonym, POS, and lemma tests. $100\%$ verbs are replaced for term tests. The words for the replacement are sampled from the STS test dataset following their frequency. For the supervised test, we sample sentence pairs with an LCS that consists of at least $80\%$ words in the shorter sentence. We use Roberta$_{base}$ as the PLM and apply Hellinger distance as the divergence function to guarantee the boundary of our metric. Mean pooling is used as the attention-assigning strategy.

\begin{equation*}
    \centering
    \begin{aligned}
    \textrm{H}(q, q') &= \frac{1}{\sqrt{2}} \sqrt{\sum_{k=1}^{c} (\sqrt{q_k}-\sqrt{q'_k})^2} \sim [0, 1]\\
    \end{aligned}
\end{equation*}

The STS experiment results are presented in Table~\ref{tab:sts}. For a fair comparison, Roberta$_{base}$ is also applied to calculate $S_C$ and PPL. NDD outperforms other metrics in all tasks, showing the strong capability of NDD to analyze semantic similarity. Also, NDD is more sensitive to POS, lemma, which is an admirable property to preserve the semantic structure for text editing. 

\begin{figure}
\begin{center}
    \ref{named}
    \centering
    \begin{tikzpicture}
    \centering
    \small
    \begin{axis}[
        legend columns=-1,
        legend entries={$\textrm{NDD}$;,$\textrm{S}_{C}$;,$\Delta \textrm{PPL}$;,$\textrm{NDD}+\textrm{S}_{C}$},
        legend to name=named,
        legend style={font=\small},
        width=8.5cm,
        height=6.0cm,
        xmin=0.0, xmax=1.0,
        ymin=-18.0, ymax=32.0,
        minor tick num=1,
        grid=both,
        grid style=dashed,
        ]
        \addplot[color=lust,mark=*,mark size=1.5pt,] coordinates{(0.0,19.0)(0.1,17.7)(0.2,15.5)(0.3,15.6)(0.4,14.9)(0.5,17.9)(0.6,18.5)(0.7,16.5)(0.8,24.0)(0.9,22.8)(1.0,30.0)};
        \addplot[color=cyan,mark=*,mark size=1.5pt,] coordinates{(0.0,26.5)(0.1,26.2)(0.2,26.2)(0.3,26.2)(0.4,25.2)(0.5, 22.4)(0.6,22.3)(0.7,19.1)(0.8,20.8)(0.9,0.0)(1.0,-10.6)};
        \addplot[color=malachite,mark=*,mark size=1.5pt,] coordinates{(0.0,-0.3)(0.1,-0.3)(0.2,-0.8)(0.3,-0.7)(0.4,-0.1)(0.5,-0.1)(0.6,0.2)(0.7,-0.4)(0.8,11.2)(0.9,-13.3)(1.0,-16.3)};
        \addplot[color=black,mark=*,mark size=1.5pt,] coordinates{(0.0,30.6)(0.1,30.3)(0.2,29.3)(0.3,29.1)(0.4,27.9)(0.5,26.9)(0.6,26.8)(0.7,22.8)(0.8,28.2)(0.9,10.3)(1.0,0.3)};
    \end{axis}
    \end{tikzpicture}
    \caption{Relationship between metric performance on the initial test and the ratio of overlapped words. The ratio $r$ in the x-axis means the evaluation to be on pairs where the overlapped rate of words in the shorter sentence of the pair $>x$.}
    \label{fig:sts}
\end{center}
\end{figure}
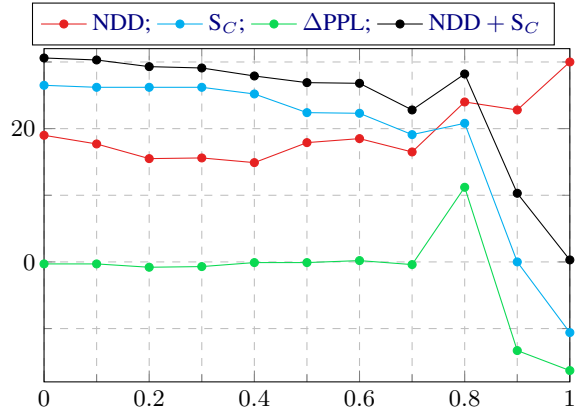

Figure~\ref{fig:sts} shows how the ratio of overlapped words affects the metric performance. $r=0$ indicates there is no overlapped word, so we are only able to use [CLS] and [SEP] tokens to evaluate the divergence. $r=1$ indicates the shorter sentence is a substring of the longer sentence as all words are overlapped. 

While S$_C$ performs better when fewer overlapped words hinder its evaluation, its performance severely suffers from a drop to even negative when the overlapped word ratio becomes $>80\%$. In contrast, the rising of the ratio helps NDD perform even better as more neighboring words participate in the evaluation to provide a precise evaluation. The ensemble (ratio = $1:0.0025$) of NDD and S$_C$ generally boosts the evaluating performance when the overlapped ratio $\leq 80\%$, indicating that NDD and S$_C$ evaluate different aspects of the semantic similarity. We further discuss the metrics using specific cases in Appendix~\ref{apdx:dist}.

\section{Unsupervised Text Compression}

The prominent performance of NDD and its correlation with overlapped word ratio inspire us to apply it for extractive text compression. Text compression takes a sentence $W$ as the input and outputs $W_C$ where $W_C$ is a substring of $W$ that maintains the main semantics in $W$. As a substring, the compressed sentence guarantees a $100\%$ overlapped ratio to support NDD's performance.

\subsection{Span Searching and Selection}

Given a sentence $W$, we try every span $W_{ij} = [w_i, \cdots, w_j]$ with length under a length limitation $\mathbb{L}_{max}$ for deletion. Then we use NDD to score the semantic difference caused by the deletions.

\begin{equation*}
    \centering
    \begin{aligned}
    W'_{ij} &= [w_1, \cdots, w_{i-1}, w_{j+1}, \cdots, w_n]\\
    NDD_{ij} &= \textrm{NDD}(W, W'_{ij})
    \end{aligned}
\end{equation*}

\begin{figure}
    \centering
    \includegraphics[width=0.5\textwidth]{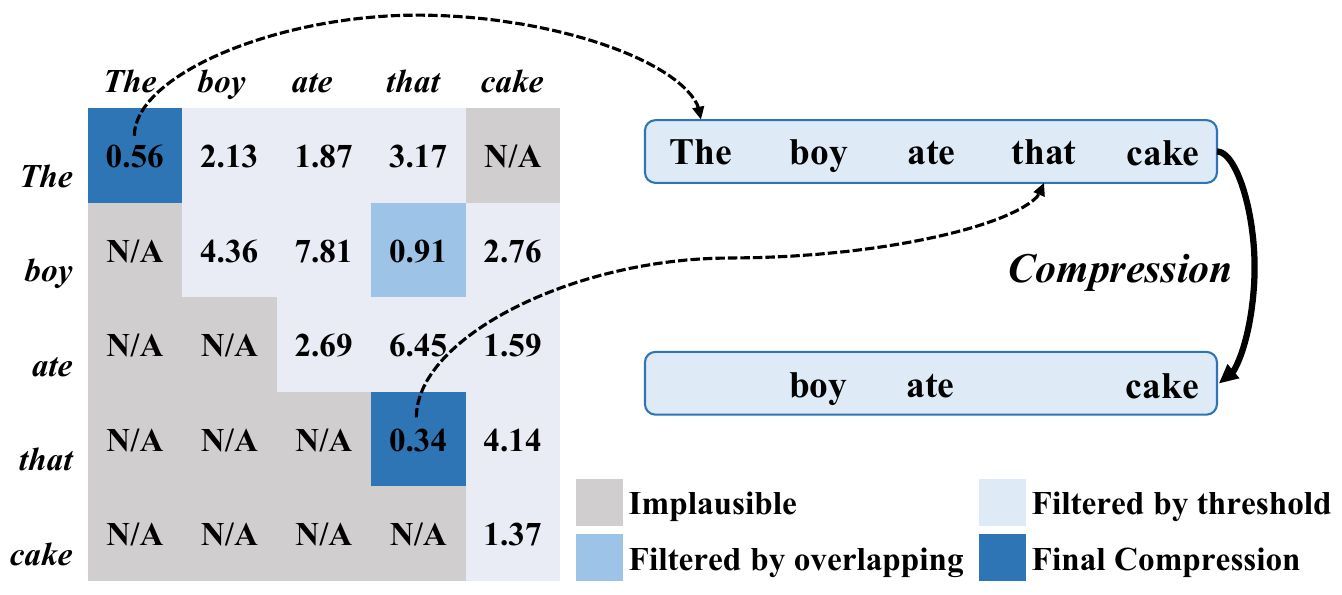}
    \caption{The compressing scenario of our NDD-based algorithm.}
    \label{fig:scenario}
\end{figure}

As in Figure~\ref{fig:scenario}, We first filter $W_{ij}$ with $NDD_{ij}$ above the threshold $\mathbb{N}_{max}$. As overlapping still exists in searched spans, we compare each overlapped span pair and drop the span with a lower NDD score. The process iterates until no overlapped candidate exists.

\begin{table}
    \centering
    \small
    \scalebox{1.0}{ 
    \begin{tabular}{llc}
    \toprule
    & Metric & Complexity \\
    \midrule
    \multirow{3}*{\rotatebox{90}{\textsc{Eval}}}& $\Delta$PPL & $2n$ \\
    & S$_C$ & $2$  \\
    & NDD & $2Len(W_{LCS})$   \\
    \midrule
    \multirow{5}*{\rotatebox{90}{\textsc{Copression}}}& PPL Deleter & $O(n^3)$ \\
    & NDD & $O(n^3)$ \\
    & NDD w/ syn. & $O(n^2)$ \\
    & Fast NDD & $O(n^2)$ \\
    & Fast NDD w/ syn. & $O(n)$ \\
    \bottomrule
    \end{tabular}
    }
    \caption{Time complexity of evaluating and compressing methods.}
    \label{tab:eff}
\end{table}

\subsection{Experiment}

\begin{table*}
    \centering
    \small
    \scalebox{1.}{
    \begin{tabular}{llccccc}
    \toprule
    & \bf \multirow{2}*{Method} & \multicolumn{1}{c}{F} & \multicolumn{3}{c}{ROUGE} & CR\\
    \cmidrule(l){3-3}
    \cmidrule(l){4-6}
    \cmidrule(l){7-6}
    & & F$_1$ & R$_1$ & R$_2$ & R$_L$ & CR \& $\Delta C$ \\
    \midrule
    & Unedited & 58.2 & 63.8 & 53.4 & 63.3 & 1.00 (+0.56) \\
    \midrule
    \multirow{5}{*}{\rotatebox{90}{\textsc{Supervised}}}& LSTM \cite{DBLP:conf/emnlp/FilippovaACKV15} & 80.0 & - & - & - & 0.39 (-0.05) \\
    & LSTM-Dep$^\ddag$ \cite{DBLP:conf/emnlp/FilippovaACKV15} & 81.0 & - & - & - & 0.38 (-0.06) \\
    & Evaluator-SLM$^\ddag$ \cite{DBLP:conf/acl/ZhaoLA18} & 85.1 & - & - & - & 0.39 (-0.05) \\
    & Tagger+BERT$^\ddag$ \cite{DBLP:conf/aaai/KamigaitoO20} & 85.0 & 78.1 & 69.9  & 77.9 & 0.40 (-0.04) \\
    & SLAHAN$^\ddag$ \cite{DBLP:conf/aaai/KamigaitoO20} & 85.5 & 79.3 & 71.4 & 79.1 & 0.42 (-0.02)   \\
    \midrule
    \multirow{9}{*}{\rotatebox{90}{\textsc{Unsupervised}}} & Drop Head & 31.7 & 23.6 & 14.7 & 22.7 & 0.39 (-0.05) \\
    & Drop Tail & 56.0 & 58.2 & \textbf{47.4} & 57.7 & 0.39 (-0.05) \\
    & Random Drop & 42.3 & 41.7 & 13.6 & 40.4 & 0.40 (-0.04) \\
    & PPL Deleter \cite{DBLP:journals/corr/abs-1909-03223} & 50.0 & -  & - & - & 0.39 (-0.05)    \\
    & PPL Deleter$^\dag$ & 50.9 & 51.3  & 36.7 & 50.9 & 0.42 (-0.02)  \\
    & NDD (Ours)& 61.2 & 60.3 & 43.2 & 59.6 & 0.41 (-0.03) \\
    & NDD+SC$^\ddag$ (Ours)& 62.3 & \bf \underline{62.6} & 45.9 & \bf \underline{61.9} & 0.42 (-0.02) \\
    & Fast NDD (Ours)& 59.7 & 55.5 & 40.3 & 54.8 & \textbf{0.43 (-0.01)} \\
    & Fast NDD+SC$^\ddag$ (Ours)& \bf \underline{67.1} & 62.0  & 46.6 & 61.4 & \textbf{0.43 (-0.01)} \\
    \bottomrule
    \end{tabular}
    }
    \caption{Results for sentence compression on the Google dataset. SC: Subtree Constraint with syntax treebanks. \underline{Underline:} the performance improvement is significant $(p<0.05)$ considering the highest baseline. $\dag$: the method is a re-implementation. $\ddag$: the method uses syntactic information.}
    \label{tab:comp}
\end{table*}

\begin{table*}
    \centering
    \small
    \scalebox{1.}{
    \begin{tabular}{lccccccc}
    \toprule
    \bf \multirow{2}*{Method} & \bf \multirow{2}*{Training} & \bf \multirow{2}*{Data} & \multicolumn{1}{c}{F} & \multicolumn{3}{c}{ROUGE} & CR\\
    \cmidrule(l){4-4}
    \cmidrule(l){5-7}
    \cmidrule(l){8-8}
    & & & F$_1$ & R$_1$ & R$_2$ & R$_L$ & CR \& $\Delta C$ \\
    \midrule
    SLAHAN$^\ddag$ & \Checkmark & \Checkmark & 57.7 & 40.1 & 30.6 & 39.6 & 0.35 (-0.36)   \\
    Fast NDD+SC$^\ddag$ & \XSolidBrush & \XSolidBrush & \bf \underline{76.5} & \bf \underline{69.8} & \bf \underline{55.1} & \bf \underline{68.5} & \textbf{0.70 (-0.01)} \\
    \bottomrule
    \end{tabular}
    }
    \caption{Comparison between the supervised state-of-the-art SLAHAN and our NDD method on BNC Corpus.}
    \label{tab:comp_bnc}
\end{table*}

\paragraph{Dataset} We conduct our experiments on two English datasets, Google dataset \cite{DBLP:conf/emnlp/FilippovaACKV15} and Broadcast News Compression (BNC) Corpus\footnote{\href{https://www.jamesclarke.net/research/resources}{https://www.jamesclarke.net/research/resources}}. On the Google dataset, we follow previous setups to use the first $1000$ sentences for testing. The BNC dataset does not have a training dataset so one of the previous works \cite{DBLP:conf/aaai/KamigaitoO20} trains a compressor on the training dataset of Google for compression. We also include a Chinese colloquial Sentence Compression (SC) dataset \footnote{\href{https://github.com/Zikangli/SOM-NCSCM}{https://github.com/Zikangli/SOM-NCSCM}} to investigate the cross-language generality of NDD. For the Chinese colloquial Sentence Compression dataset, we replaced the masks of entities with their natural language expressions\footnote{Can be found in Appendix~\ref{apdx:sub}} to avoid inaccuracy caused by them to NDD calculation.

\paragraph{Configuration} We take cased BERT$_{base}$ as the PLM for English and BERT$_{Chinese}$ for Chinese. The divergence function is set to 
Kullback–Leibler (KL) divergence. The prediction on the initial text is used as the approximating distribution since it is predicted based on the text with an integral structure.  

\begin{equation*}
    \centering
    \begin{aligned}
    \textrm{D}_{KL}(q, q') &= \sum_{k=1}^{c} q'_k\log(\frac{q'_k}{q_k}) \sim [0, \infty)\\
    \end{aligned}
\end{equation*}

We fix the following hyperparameters during the experiment. $\mathbb{L}_{max}$ is set to $9$ when the syntax is used and else $5$. $\mathbb{N}_{max}$ is set to $1.0$. Our compression is iterated for at most $5$ times until no word is deleted. The weighing process can be referred to Appendix. Other parameters are adjusted to control the compression ratio. Considering the time complexity of NDD, we have developed a faster variant called Fast NDD. Fast NDD calculates the divergence by considering only the two adjacent words of the compressed span. This approach is based on the hypothesis that the nearest words are the most affected by span switching. As the effect of syntax information is shown to be effective in supervised text compression \cite{DBLP:conf/aaai/KamigaitoO20}, we add a constraint that only allows dropped spans to subtrees in the syntactic dependency treebank for each step. This also boosts the efficiency as we only need to consider sparse subtree spans. The efficiency of different scenarios of NDD is shown in Table~\ref{tab:eff}. Here the time complexity refers to the times of PLM-based MLM or presentation calculation. $n$ and $k$ refer to the length of the sentence and the dropped span, respectively.

\paragraph{Metric} We apply the commonly-used F1 score and ROUGE metric \cite{lin-2004-rouge} to evaluate the overlapping between our compression and the golden one and compare with previous works. For ROUGE, we follow the evaluating scenario in \cite{DBLP:conf/aaai/KamigaitoO20} to truncate the parts in the prediction that exceed the byte length of the golden one. We also incorporate BLEU \cite{papineni-etal-2002-bleu} to compare with baselines that report BLEU on the Chinese colloquial Sentence Compression dataset. Compression ratio (CR) refers to the percentage of preserved sentences in the initial sentence and $\Delta C = \textrm{CR}_{pred} - \textrm{CR}_{gold}$, which is better when being closer to $0$. 

\paragraph{Baseline} We use the PPL Deleter \cite{DBLP:journals/corr/abs-1909-03223} as the main baseline. Deleter uses $\Delta$PPL to control the compressing procedure and tries to preserve a lower PPL in each step. Simple baselines that directly drop words according to the compression ratio are also included. We report several supervised results to show the current development on the tasks.

\paragraph{Google} Table~\ref{tab:comp} presents our results on the Google dataset. Compared to the PPL Deleter, the basic NDD leads to a sharp improvement on all metrics, $10.3$ improvement on the F1 score, and $8.7$ on ROUGE$_L$. Fast NDD underperforms the initial NDD, but the performance is admirable considering its efficiency. Our method benefits from syntactic constraints, especially for Fast NDD. Syntactic constraints boost Fast NDD's performance to around $7.0$ on most metrics to set the new unsupervised state-of-the-art on F1 score. Still, the initial NDD method with syntax is state-of-the-art on ROUGE metrics. Thanks to the compression rate controlling ability of our method, we can control the compression to a CR extremely close to the golden one.

\paragraph{BNC} The BNC Corpus is a perfect case to show the advantage of NDD's ability to control the compression rate. We take the supervised SOTA syntactically look-ahead attention network (SLAHAN) \cite{DBLP:conf/aaai/KamigaitoO20} as the baseline. Since BNC does not have a training dataset, SLAHAN is trained on the $200K$ Google corpus. Nevertheless, the cross-domain adaption of SLAHAN is not successful as its $\Delta C$ is an extremely negative $-0.35$ in Table~\ref{tab:comp_bnc}. In contrast, our PLM-based unsupervised method enjoys robustness and can be easily adapted to different domains, and reach a CR close to the golden one. Our unsupervised method thus outperforms the supervised state-of-the-art by a huge margin ($20\sim 30$) on all metrics. 

\begin{table*}
    \centering
    \small
    \begin{tabular}{llccccccccc}
    \toprule
    & \bf \multirow{2}*{Method} & \multicolumn{1}{c}{F} & \multicolumn{4}{c}{BLEU} & \multicolumn{3}{c}{ROUGE} & CR\\
    \cmidrule(l){3-3}
    \cmidrule(l){4-7}
    \cmidrule(l){8-10}
    \cmidrule(l){11-11}
    & & F$_1$ & B$_1$ & B$_2$ & B$_3$ & B$_4$ & R$_1$ & R$_2$ & R$_L$ & CR \& $\Delta C$ \\
    \midrule
    & Unedited & 81.7 & 74.4 & 73.7 & 71.4 & 68.5 & 70.4 & 55.3 & 70.2 & 1.00 (+0.32)    \\
    \midrule
    \multirow{2}{*}{\rotatebox{90}{\textsc{Sup.}}} & Tagger+BERT \cite{DBLP:conf/emnlp/ZiWLLCC21} &  85.8 & 73.4 & 64.3  & 58.3 & 53.9 & - & -  & - & 0.83 (+0.15) \\
    & SOM-NCSCM \cite{DBLP:conf/emnlp/ZiWLLCC21} &  89.7 & 84.0 & 78.2  & 74.8 & 70.1 & - & -  & - & \textbf{0.68 (-0.00)} \\
    \midrule
    \multirow{7}{*}{\rotatebox{90}{\textsc{Unsupervised}}} & PPL Deleter$^\dag$ & 56.3 & 51.8 & 47.3  & 43.5 & 40.5 & 55.7  & 25.3 & 55.4 & 0.70 (+0.02)  \\
    & NDD (Ours)& 74.8 & 66.4 & 57.1 & 50.5 & 44.6 & 67.8 & 47.0 & 67.8 & 0.65 (-0.03) \\
    & NDD+SC$^\ddag$ (Ours) & 76.4 & 68.5 & 59.3 & 52.8 & 47.2 & 69.3 & 49.9 & 69.2 & \textbf{0.68 (-0.00)} \\
    & NDD (wwm)+SC$^\ddag$ (Ours)& \textbf{\underline{76.7}} & \textbf{\underline{68.6}} & \textbf{\underline{59.6}} & \textbf{\underline{53.5}} & \textbf{\underline{47.9}} & \textbf{\underline{70.5}} & \textbf{\underline{50.8}} & \textbf{\underline{70.2}} & 0.70 (+0.02) \\
    & Fast NDD (Ours)& 73.9 & 64.1 & 56.6 & 51.2 & 45.6 & 65.9 & 48.4 & 65.7 & 0.67 (-0.02) \\
    & Fast NDD+SC$^\ddag$ (Ours)& 75.5 & 66.9 & 58.5 & 52.4 & 47.2 & 67.2 & 47.7 & 66.8 & 0.69 (+0.01) \\
    & Fast NDD (wwm)+SC$^\ddag$ (Ours)& 74.7 & 66.0 & 57.7 & 51.7 & 45.7 & 67.7 & 48.4 & 67.5 & 0.70 (+0.02) \\
    \bottomrule
    \end{tabular}
    \caption{Results for sentence compression on the Chinese colloquial Sentence Compression dataset.}
    \label{tab:comp_zh}
\end{table*}

\paragraph{Colloquial SC} The experimental results depicted in Table~\ref{tab:comp_zh} underline the cross-lingual generality of our NDD method. Notably, our method surpasses the PPL Deleter in performance, thereby setting a new benchmark for unsupervised models across all evaluation metrics. The compression rate controlling ability of NDD further allows it to generate BLEU scores that are in close alignment with the supervised Tagger+BERT model, indicating the strength of our approach. In our experiments, we also deployed a whole-word-masking (wwm) Roberta\footnote{\href{https://huggingface.co/hfl/chinese-roberta-wwm-ext}{https://huggingface.co/hfl/chinese-roberta-wwm-ext}} as the PLM. This led to additional performance enhancements, which indicates the accuracy of NDD can benefit from whole-word-masking during the pre-training.

In summary, our NDD method coupled with the Subtree Constraint offers the best overall performance among unsupervised models. It achieves the highest F$_1$ score of $76.7$, surpasses all others in most metrics, and is very close to the best CR. This confirms its strong potential for the task of sentence compression across languages.

\begin{table}
    \centering
    \small
    \scalebox{.95}{
    \begin{tabular}{lccccc}
    \toprule
    \multirow{2}*{$\mathbb{N}_{max}$/$\mathbb{L}_{max}$} & \multicolumn{1}{c}{F} & \multicolumn{3}{c}{ROUGE} & CR\\
    \cmidrule(l){2-2}
    \cmidrule(l){3-5}
    \cmidrule(l){6-6}
     & F$_1$ & R$_1$ & R$_2$ & R$_L$ & CR \& $\Delta C$ \\
    \midrule
    4.00 / 5 & 44.6 & 35.8 & 15.4 & 35.7 & 0.16 (-0.28) \\
    2.00 / 5 & 58.5 & 52.4 & 34.9 & 52.1 & 0.27 (-0.17) \\
    1.00 / 5 & 67.1 & 62.0  & 46.6 & 61.4 & \textbf{0.43 (-0.01)} \\
    0.50 / 5 & \bf 67.7 & 64.6 & 51.2 & 64.1 & 0.57 (+0.13) \\
    0.25 / 5 & 66.0 & \bf 65.1 & \bf 53.0 & \bf 64.6 & 0.69 (+0.25) \\
    1.00 / 2 & 66.8 & 63.0 & 47.5 & 62.4 & 0.48 (+0.04) \\
    1.00 / 1 & 64.7 & 64.2 & 49.3 & 63.7 & 0.62 (+0.18) \\
    \bottomrule
    \end{tabular}
    }
    \caption{Performance results of different configuration setups on the Google dataset.}
    \label{tab:var}
\end{table}

\paragraph{Compression Rate Controlling} We provide a more specific analysis of NDD's compression rate controlling ability. By changing the configuration of our scenario, our method can result in different compression ratios, from $16\%$ to $69\%$. When the compression ratio is higher than $43\%$, NDD always results in text with admirable quality (F1 $> 60\%$, ROUGE$_{1\&L} > 60\%$). Also, when the CR is extremely small, NDD can still preserve much information in the initial sentence, with overlapping F1 score $58.5$ for $27\%$ and $44.6$ for $16\%$. Also, adjusting the compressing iteration for the same configuration setup can result in high-quality output in different compression ratios. The compression rate controlling ability enables our method to easily adapt to systems requiring different compression ratios. Further case-based discussion can be referred to Appendix~\ref{apdx:case}.

\section{Further Analysis\footnote{In analyses, we continue using the KL divergence as the divergence function.}}

We continue studying the compression algorithm to further investigate NDD's syntax awareness via analyzing the roles of pruned words in the syntax treebank.

\subsection{Syntax Subtree Pruning}

\begin{figure}
    \centering
    \includegraphics[width=0.5\textwidth]{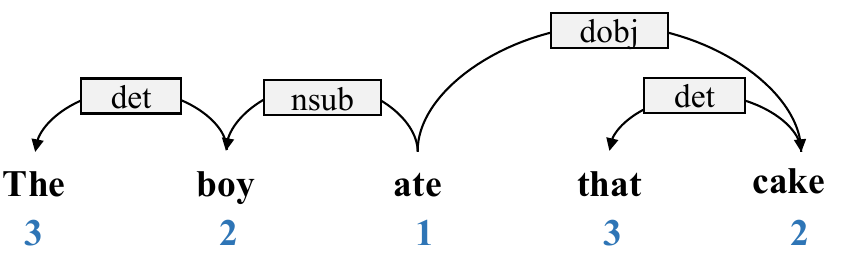}
    \caption{Nodes and their depths (blue) in a syntax treebank. Deeper nodes in the treebank generally play less important roles in context.}
    \label{fig:tree}
\end{figure}

 This task tests whether NDD is able to detect syntactic structures using syntax treebanks. (1) If the pruned nodes mostly play subordinated roles in the tree, our algorithm can be better certificated to compress with an awareness of syntax. We depict an instance of syntax treebank in Figure~\ref{fig:tree}. In the treebank, deeper nodes like \textit{the} and \textit{that} are less important for the integrity of syntax structure. (2) Also, pruning a subtree like \textit{that cake} will preserve more syntax structure than pruning a non-subtree like \textit{ate that}. Thus, we introduce two metrics to evaluate the pruning performance: \textbf{Depth-$n$} and \textbf{Subtree-$k$}.

\begin{equation*}
    \centering
    \begin{aligned}
    \textrm{Depth-}n &= \frac{Count(w|Depth(w) = n)}{Count(w)}\\
    \textrm{Subtree-}k &= \frac{Count(s|IsSub(s), Len(s)=k)}{Count(s|Len(s)=k)}\\
    \end{aligned}
\end{equation*}

\noindent where $w$, $s$ represent the pruned words and spans. $Count(\cdot)$ returns the number of items, $Depth(\cdot)$ returns the depth of a word in the syntax treebank, $IsSub(\cdot)$ return if a span is a subtree in the treebank, and $Len(\cdot)$ returns the number of words in a span. Depth-$n$ and Subtree-$k$ thus reflect the word-level and span-level pruning quality, respectively. 

We experiment on the PTB-3.0 test dataset \cite{DBLP:journals/coling/MarcusSM94}. We use the random dropping strategy with the same compression ratio as the baseline for comparison. As in Table~\ref{tab:syn}, the proportion of nodes in shallower levels (depth=$1\sim3$) pruned by our algorithm is smaller than all the corresponding random and PPL-based pruning. Also, the proportion of subtrees in spans pruned by the NDD-based algorithm is significantly larger than in other correspondents. Thus, we conclude that NDD can guide the compressing algorithm to detect subordinated components in syntax dependency treebanks.

\begin{table}
    \centering
    \scalebox{0.83}{ 
    \begin{tabular}{lcccccccc}
    \toprule
    \bf \multirow{2}*{Method} & \multirow{2}*{$\mathbb{N}_{max}$} & \multicolumn{4}{c}{\textbf{Depth-}$n$} & \multicolumn{3}{c}{\textbf{Subtree-}$k$} \\
    \cmidrule(l){3-6}
    \cmidrule(l){7-9}
    & & 1 & 2 & 3 & $\geq$ 4 & 1 & 2 & $\geq$ 3 \\
    \midrule
    \multirow{2}*{Random} & 1.0 & 4 & 21 & 22  & 54 & 55 & 31 & 27  \\
    & 2.0 & 4 & 24 & 23  & 48 & 52 & 29 & 23  \\
    \midrule
    \multirow{2}*{PPL} & 1.0 & 3 & 20 & 21  & 56 & 79 & 57 & 48  \\
    & 2.0 & 3 & 23 & 24  & 50 & 71 & 45 & 37  \\
    \midrule
    \multirow{2}*{NDD} & 1.0 & 1 & 19 & 20  & 60 & 90 & 81 & 66  \\
    & 2.0 & 2 & 23 & 23  & 52 & 82 & 70 & 63  \\
    \bottomrule
    \end{tabular}
    }
    \caption{Proportion (\%) of pruned nodes in certain depths of the syntax treebanks and proportion (\%) of pruned spans that are subtrees. $\mathbb{L}_{max}$ is set to $5$.}
    \label{tab:syn}
\end{table}

\subsection{Predicate Detection}

To explore the semantics awareness of NDD, we experiment on the semantic role labeling (SRL) task for predicate detection. As predicates are semantically related to more components (augments) in sentences, deleting them or replacing them with stop words will result in a larger semantic distance from the initial sentence. We evaluate the predicate detecting ability following the words ranking task. We rank the probability of words to be predicates according to NDD evaluation and evaluate the detecting performance by ranking metrics: mean average precision (mAP) and area under curve (AUC).

We conduct our experiments on Conll-2009 SRL datasets\footnote{Instances with length $\leq 50$, number of predicates $>0$.} \cite{DBLP:conf/conll/HajicCJKMMMNPSSSXZ09}. To test our method‘s generality, in-domain (ID) and out-of-domain (OOD) English (ENG) datasets are included. Another Spanish (SPA) dataset is also used for cross-language evaluation. To generate a new sentence for semantic distance computation, we edit each word in the sentence in three ways: (a) Deletion, (b) Replacement with a mask token, (c) Replacement with a stop word\footnote{\textit{a} for ENG-ID, \textit{that} for ENG-OOD and \textit{el} for SPA}. We apply cased SpanBERT$_{base}$ \cite{DBLP:journals/tacl/JoshiCLWZL20} and cased BERT$_{Spanish}$\footnote{\href{https://huggingface.co/dccuchile/bert-base-spanish-wwm-cased}{https://huggingface.co/dccuchile/bert-base-spanish-wwm-cased}} \cite{CaneteCFP2020} as PLMs. For comparison, we implement a PPL-based algorithm that uses $\Delta$PPL to detect predicates. 

Our results are presented in Table~\ref{tab:sem}. The generally poor performance shows that $\Delta$PPL might not be a proper metric for predicate detection. In contrast, the NDD-based algorithm produces much better results and outperforms the PPL-based algorithm by $10\sim 20$ scores on both AUC and mAP metrics, which is a remarkably significant margin and verifies NDD to be much more capable in understanding semantics. The ensemble of three processes boosts AUC, mAP to higher than $80.0$, $50.0$, respectively, making it a plausible way to detect predicates following an unsupervised procedure. 

\begin{table}
    \centering
    \scalebox{0.72}{ 
    \begin{tabular}{lcccccc}
    \toprule
    \bf \multirow{2}*{Edit}& \multicolumn{2}{c}{ENG-ID}& \multicolumn{2}{c}{ENG-OOD}& \multicolumn{2}{c}{SPA}\\
    \cmidrule(l){2-3}
    \cmidrule(l){4-5}
    \cmidrule(l){6-7}
    & mAP & AUC & mAP & AUC & mAP & AUC  \\
    \midrule
    \textit{(PPL-based)}\\
    Delete & 36.8 & 56.8 & 44.5 & 60.4 & 26.6 & 54.5     \\
    Mask Replace & 35.9 & 56.7 & 33.1 & 48.5 & 25.1 & 50.4    \\
    \midrule
    \textit{(NDD-based)}\\
    Delete & 53.1 & 77.0 & 62.5 & 80.8 & 48.8 & 77.1   \\
    Mask Replace & 48.0 & 74.4 & 57.3 & 80.6 & 44.2 & 75.0   \\
    Stop Word Replace & 49.9 & 76.4 & 56.4 & 78.5 & 45.2 & 77.2   \\
    Ensembled & \bf \underline{54.3} & \bf \underline{79.7} & \bf \underline{63.1} & \bf \underline{83.3} & \bf \underline{54.7} & \bf \underline{82.8}   \\
    \bottomrule
    \end{tabular}
    }
    \caption{Evaluation on ability of metrics to detect predicates in sentences.}
    \label{tab:sem}
\end{table}

\section{Related Works}

The evaluation on text similarity provides valuable guidance on various downstream tasks, including text classification \citep{DBLP:journals/aai/ParkHK20}, document clustering \cite{DBLP:journals/ijbidm/LakshmiB21}, and translated text detection \cite{DBLP:conf/naacl/Nguyen-SonPHGK21}. The commonly used cosine similarity evaluates paired sentences' similarity based on the cosine value between word embeddings or pre-trained representations \cite{reimers-gurevych-2019-sentence,zhang-etal-2020-unsupervised}. Unfortunately, when the overlapping ratio between paired sentences rises, the representation-based method suffers from faults caused by similar word representations. Our work replaces word representations with predicted distributions to mitigate the disturbance from overlapped components. 

The proposal of PLMs \cite{DBLP:conf/naacl/DevlinCLT19} inspires researchers to leverage the upstream training process for text similarity evaluation. \citeauthor{DBLP:journals/corr/abs-1909-03223} leverage the perplexity calculated from PLMs to represent the semantic distance between texts during text compression. While perplexity can evaluate the fluency of sentences, a recent study \cite{DBLP:conf/acl/KuribayashiOIYA20} suggests that low perplexity does not directly refer to a human-like sentence. Also, perplexity fails with words that share a similar existing probability but are with opposite or irrelevant meanings. Other PLM-based metrics like BERTScore have been verified by experiments to evaluate text generation better \cite{DBLP:conf/iclr/ZhangKWWA20}. Other pre-trained models for evaluation are also an interesting topic. To evaluate semantics preservation in AMR-to-sentence, \citeauthor{DBLP:conf/eacl/OpitzF21} exploits AMR parser to compare the AMR graph of generated results with the golden graph, showing the potential of pre-trained models to evaluate more complex linguistic structures.

Many supervised methods \citep{DBLP:conf/acl/MalireddyMS20,DBLP:conf/propor/NobregaJBP20} have been proposed for text compression. Syntax treebanks play a critical role in text compression \cite{xu-durrett-2019-neural,DBLP:conf/cncl/WangC19,DBLP:conf/aaai/KamigaitoO20}. Unsupervised methods have been explored to extract sentences from documents to represent key points \cite{DBLP:journals/access/JangK21}. Nevertheless, span pruning is still far from satisfaction. As mentioned before, \cite{DBLP:journals/corr/abs-1909-03223} explores using $\Delta$PPL for compression, which is not so capable as NDD in semantics preservation. 

Syntax and semantic analyses \cite{DBLP:conf/iclr/DozatM17,DBLP:conf/aaai/LiZP20,DBLP:conf/acl/LiJPS20,DBLP:conf/emnlp/LiZWP20} reflect model's awareness of the internal structures in sentences. The awareness of syntax and semantics of NDD is verified by those tasks.

\section{Conclusion}

We address the overlapping issue in semantic distance evaluation in this paper. To mitigate the disturbance from overlapped components, we mask and predict words in the LCS via PLM-based MLM. NDD evaluates the semantic distance using a weighted sum of the divergence between predicted distributions. STS experiments verify NDD to be more sensitive to a wide range of semantic differences and perform better on highly overlapped paired texts, which is challenging for conventional metrics. NDD-based text compression algorithm significantly boosts the unsupervised performance, and its high compression rate controlling ability enables the adaption to datasets in different domains. NDD's awareness of syntax and semantics is verified by further analyses, showing the potential of NDD for further studies. 

\section*{Limitations}

While our NDD metric has demonstrated its effectiveness in measuring the semantic distance between overlapped sentences, there are still some limitations to consider. Firstly, the calculation efficiency of NDD may become a bottleneck when dealing with large amounts of data. The mask-and-predict strategy requires the generation of a large number of predictions for each word in the LCS, which can be computationally expensive. Therefore, for large-scale applications, more efficient algorithms or hardware acceleration may be necessary to speed up the calculation of NDD. Secondly, our method currently cannot selectively compress certain parts of the text. The mask-and-predict strategy compresses the entire overlapped segment, which may not always be desirable. For example, in some cases, it may be more desirable to compress only the less relevant portion of the text while retaining the most informative content. While NDD has an advantage over supervised compressors in controlling compression ratio, it still cannot control the compression orders. Future research may investigate techniques to allow for more fine-grained control over the compression process. Overall, while NDD shows great promise in improving the evaluation of semantic similarity and text compression, further research is needed to address these limitations and improve the compression rate controlling ability and versatility of the method.

\bibliography{anthology,custom}
\bibliographystyle{acl_natbib}

\clearpage

\appendix

\section{Dataset Statistics}

\begin{table}[H]
    \centering
    
    \scalebox{0.85}{ 
    \begin{tabular}{lccc}
    \toprule
    Dataset & Inst. Num. & Avg. Len. & CR \\
    \midrule
    Google & 1000 & 27.71 & 0.44 \\
    BNC & 595 & 31.15 & 0.71 \\
    Colloquial SC & 150 & 8.13 & 0.68 \\
    \midrule
    STS-B & 2758 & 9.81 & - \\
    PTB & 2416 & 23.46 & - \\
    Conll09-ENG-ID & 2399 & 24.04 & - \\
    Conll09-ENG-OOD & 425 & 16.96 & - \\
    Conll09-SPA & 1725 & 29.35 & - \\
    \bottomrule
    \end{tabular}
    }
    \caption{Statistics of our datasets in experiments.}
    \label{tab:stat}
\end{table}

\section{Specific Cases for Semantic Difference Evaluation}
\label{apdx:dist}

We use specific cases to further explore the ability of NDD to capture precise semantic differences using several examples. As in Table~\ref{tab:pss}, we edit the initial sentence \textit{"I am walking in the cold rain."} with a series of replacements. We keep the syntactic structure of the sentence unchanged and replace some words with other words of the same part-of-speech. Thus, the difference between the initial and edited sentences is majorly the semantics.

\begin{table}[H]
    \centering
    \small
    \scalebox{0.85}{ 
    \begin{tabular}{p{5cm}ccc}
    \toprule
    Sentence & PPL & NDD & S$_C-$\\
    \midrule
    I am walking in the cold rain. & 5.99 & 0.00 & 1.000\\
    \midrule
    I am walking in the \underline{cool} rain. & 10.10 & 0.81 & 0.995 \\
    I am walking in the \underline{freezing} rain. & 5.63 & 0.97 & 0.997 \\
    I am walking in the \underline{heavy} rain. & 5.30 & 1.82 & 0.994 \\
    I am walking in the \underline{hot} rain. & 14.77 & 3.17 & 0.995 \\
    \midrule
    I am walking in the cold \underline{snow}. & 5.37 & 2.46 & 0.996 \\
    I am walking in the cold \underline{night}. & 6.18 & 3.52 & 0.991 \\
    I am walking in the cold \underline{sunshine}. & 8.59 & 4.73 & 0.994 \\
    \midrule
    I am \underline{running} in the cold rain. & 11.86 & 0.66 & 0.990\\
    I am \underline{wandering} in the cold rain. & 16.89 & 0.89 & 0.982\\
    I am \underline{swimming} in the cold rain. & 14.84 & 3.29 & 0.986\\
    \midrule
    I \underline{was} walking in the cold rain. & 10.32 & 4.72 & 0.980\\
    \underline{He} am walking in the cold rain. & 105.55 & 13.04 & 0.991\\
    \underline{He is} walking in the cold rain. & 13.95 & 7.22 & 0.980\\
    \bottomrule
    \end{tabular}
    }
    \caption{Cases for detection of NDD on very precise semantic difference. The initial sentence is \textit{"I am walking in the cold rain."}}
    \label{tab:pss}
\end{table}

We divide the editing cases into several groups. In the first three groups, we change words (adjective, noun, and verb respectively) into similar, different, or opposite meanings. NDD successfully detects the semantic difference and precisely evaluates changing extents. Taking the first group as an instance, changing from \textit{cold} into \textit{cool} and \textit{freezing} keeps most semantics while changing into \textit{hot} leads to the opposite and even implausible semantics. NDD reflects the difference of semantics between these edited results and assigns a much higher score to the \textit{cold}-to-\textit{hot} case. Moreover, in the medium case where the aspect for description is changed to \textit{heavy}, NDD remarkably assigns a medium score to this case, showing its high discerning capability.

In the last case group, we change the tense and subject of the sentence. NDD is shown to be fairly sensitive to tenses and subjects. This property can be used to retain those critical properties during edits. NDD is also able to detect syntactic faults like the combination of \textit{He am} and can thus be used for fault prevention during the edit.

From these cases, we can also see why perplexity and cosine similarity is incapable of detecting precise semantic difference as NDD. In Table~\ref{tab:pss}, cosine similarity cannot detect the subtle semantic difference and even syntactic faults. We attribute this to the high reliance on word representations for sentence representations, as sentences with many words overlapped will be classified to be similar. 

For perplexity (PPL), the first problem with it is that this metric evaluates the fluency of a single sentence. Perplexity will thus guide edits to transform sentences into more syntactically plausible versions, ignoring semantics. As a result, edited results with lower perplexity may change semantics like \textit{cold}-to-\textit{heavy} and \textit{rain}-to-\textit{snow}. NDD is able to preserve semantics much better by suggesting changing \textit{cold} to \textit{cool} or \textit{freezing} and changing \textit{walking} to \textit{running} or \textit{wandering}.

Another reason is that perplexity can easily be misguided by low-frequency words. In the \textit{walking}-to-\textit{wandering} case, since \textit{wandering} is a low-frequency word, the resulted perplexity is even higher than the \textit{walking}-to-\textit{swimming} case. Since perplexity is scored based on the existence probability of words, the low-frequency \textit{wandering} will lead to a higher perplexity, even though \textit{wandering} is semantically closer to \textit{walking} than \textit{swimming}. This issue is overcome in NDD as we use predicted distributions rather than real words. As described before, NDD can understand low-frequency words and even named entities much better. As a result, NDD correctly scores the semantic difference caused by replacement on \textit{walking}.

\section{Other Details for Compression}

For weighing in text compression, we modify the exponential weight and use the balanced weights for distance.

\begin{equation*}
    \centering
    \begin{aligned}
        a_k &= \mu^{min(|k-i|, |k-j|)}\\
        a'_k &= a_k + a_{n'-k} * \mu^{n'}\\
        n' &= n - (j - i + 1) \\
    \end{aligned}
\end{equation*}

\noindent where $n$ is the length of the initial sentence, $k$ is the neighboring word's position, and $i$, $j$ are the start and end positions of the pruned span. The modification guarantees the total distance weights are the same for each NDD calculation, while the exponential weight assigns fewer weights to words on two sides of the sentence. 

\begin{figure}
    \centering
    \includegraphics[width=0.5\textwidth]{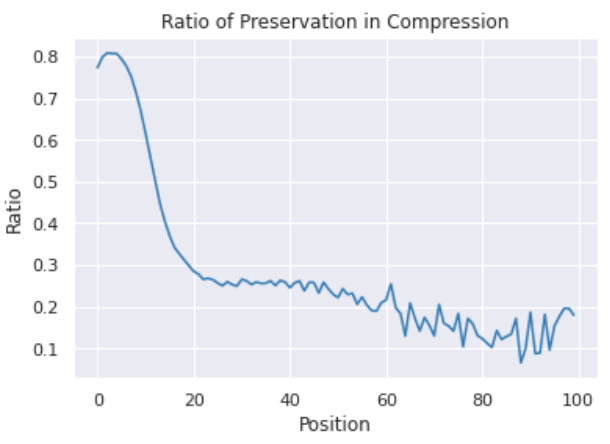}
    \caption{The ratio of preserved tokens in certain positions of the initial sentence. Statistics from the Google training dataset.}
    \label{fig:distance}
\end{figure}

Furthermore, we add another weight $b_k$ to encourage our algorithm to delete later words in the sentence. As shown in Figure~\ref{fig:distance}, later words are less common to be used for summary. We modify the weighted sum as follows.

\begin{equation*}
    \centering
    \begin{aligned}
        b_w &= \nu^{Idx(w)}\\
    NDD &= \sum_{w \in W_{LCS}} a'_w b'_w\text{F}_{div}(q_{\textrm{Idx}^d(w)}, q'_{\textrm{Idx}'^d(w)})
    \end{aligned}
\end{equation*}

In experiments, we fix $\mu$ to $0.9$ and adjust $\nu$ to adapt to the compression rate.

\section{Mask to Expression}
\label{apdx:sub}

\begin{table}[H]
    \centering
    
    \scalebox{1.}{ 
    \begin{tabular}{ll}
    \toprule
    Mask & Expression \\
    \midrule
    \begin{CJK}{UTF8}{gbsn} [业务名词] \end{CJK} & \begin{CJK}{UTF8}{gbsn} 某业务 \end{CJK}\\
    \begin{CJK}{UTF8}{gbsn} [电话号码] \end{CJK} & \begin{CJK}{UTF8}{gbsn} 电话 \end{CJK}\\
    \begin{CJK}{UTF8}{gbsn} [编号] \end{CJK} & \begin{CJK}{UTF8}{gbsn} 1 \end{CJK}\\
    \begin{CJK}{UTF8}{gbsn} [名词] \end{CJK} & \begin{CJK}{UTF8}{gbsn} 这个 \end{CJK}\\
    \begin{CJK}{UTF8}{gbsn} [地址] \end{CJK} & \begin{CJK}{UTF8}{gbsn} 某地 \end{CJK}\\
    \bottomrule
    \end{tabular}
    }
    \caption{The dictionary that transforms Chinese masks to natural language expressions.}
    \label{tab:sub}
\end{table}

\section{Extra Comparison}

\begin{table}
    \centering
    \small
    \scalebox{0.9}{
    \begin{tabular}{lccccc}
    \toprule
    \bf \multirow{2}*{Method} & \multicolumn{1}{c}{F} & \multicolumn{3}{c}{ROUGE} & CR\\
    \cmidrule(l){2-2}
    \cmidrule(l){3-5}
    \cmidrule(l){6-6}
    & F$_1$ & R$_1$ & R$_2$ & R$_L$ & CR \& $\Delta C$ \\
    \midrule
    PPL Deleter & 50.9 & 51.3  & 36.7 & 50.9 & 0.42 (-0.02)  \\
    PPL Deleter+SC$^\ddag$ & 53.1 & 54.7  & 40.3 & 54.5 & 0.42 (-0.02)  \\
    S$_C$ & 46.5 & 45.0 & 16.9 & 43.9 & 0.37 (-0.07) \\
    S$_C$+SC$^\ddag$ & 49.5 & 50.5 & 23.0 & 49.7 & 0.42 (-0.02) \\
    BERTScore & 48.0 & 48.6 & 15.8 & 47.4 & 0.41 (-0.03) \\
    BERTScore+SC$^\ddag$ & 47.2 & 49.5 & 18.8 & 48.6 & 0.39 (-0.05) \\
    NDD$^*$& 60.1 & 59.8 & 41.7 & 59.2 & 0.41 (-0.03) \\
    NDD+SC$^{*\ddag}$& 60.8 & 61.9 & 44.5 & 61.3 & \textbf{0.45 (+0.01)} \\
    NDD& 61.2 & 60.3 & 43.2 & 59.6 & 0.41 (-0.03) \\
    NDD+SC$^\ddag$& 62.3 & \bf {62.6} & 45.9 & \bf {61.9} & 0.42 (-0.02) \\
    Fast NDD& 59.7 & 55.5 & 40.3 & 54.8 & \textbf{0.43 (-0.01)} \\
    Fast NDD+SC$^\ddag$& \bf {67.1} & 62.0  & \bf 46.6 & 61.4 & \textbf{0.43 (-0.01)} \\
    \bottomrule
    \end{tabular}
    }
    \caption{Extra comparison including S$_C$ and BERTScore. $*$: use $kl(d||d')$ instead of $kl(d'||d)$}
    \label{tab:comp_extra}
\end{table}

We further investigate the capability difference of different metrics in text compression. As in Table~\ref{tab:comp_extra}, we replace the evaluator in the compressing scenario with S$_C$ and BERTScore \cite{DBLP:conf/iclr/ZhangKWWA20}. The experiment results show a large gap between NDD and other metrics, verifying the prominent semantic distance evaluating the capability of NDD. 

\section{Human Evaluation}

\begin{table}
    \centering
    \scalebox{1.}{
    \begin{tabular}{lcc}
    \toprule
    Method & Syntax & Semantics\\
    \midrule
    PPL Deleter & 3.69 & 2.56 \\
    PPL Deleter+SC$^\ddag$ & 3.93 & 2.88 \\
    S$_C$ & 1.87 & 1.51 \\
    S$_C$+SC$^\ddag$ & 2.35 & 2.08 \\
    BERTScore & 1.96 & 1.83 \\
    BERTScore+SC$^\ddag$ & 2.38 & 2.17 \\
    NDD & 3.87 & 3.46 \\
    NDD+SC$^\ddag$ & \bf 4.08 & \bf 3.62 \\
    Fast NDD & 3.73 & 3.18 \\
    Fast NDD+SC$^\ddag$ & 4.01 & 3.43 \\
    \bottomrule
    \end{tabular}
    }
    \caption{Human evaluation on the syntax and semantics integrity of outputs from unsupervised text compression algorithms.}
    \label{tab:comp_huamn}
\end{table}

We further use human evaluation to compare the performance of text compression algorithms. We sample $100$ sentences from the Google test dataset and ask human evaluators to score for the syntactic and semantic integrity of the output. The evaluators are blind to which algorithm compresses and produces the output. We assign scores from $0$ to $5$ as follows:

\begin{itemize}
    \item $0$: No legal structure, totally a combination of meaningless fragments.
    \item $1$: Poor structure, only some meaningful components, and the whole structure are not understandable. 
    \item $2$: The whole structure is acceptable but contains faults compared to the initial sentence. 
    \item $3$: Some parts of the initial structure are preserved, but the compression drops some important components. 
    \item $4$: Most parts of the initial structure are preserved, still there exists a little inconsistency or ignorance of important components.
    \item $5$: The structure is as integral as human's. 
\end{itemize}

The human evaluation verifies NDD to keep a large gap with conventional metrics in text compression in syntactic and semantic integrity. Also, the benefit of introducing syntactic constraints is shown in every algorithm. 

\section{NDD distributions}

\begin{figure*}
    \centering
    \includegraphics[width=0.99\textwidth]{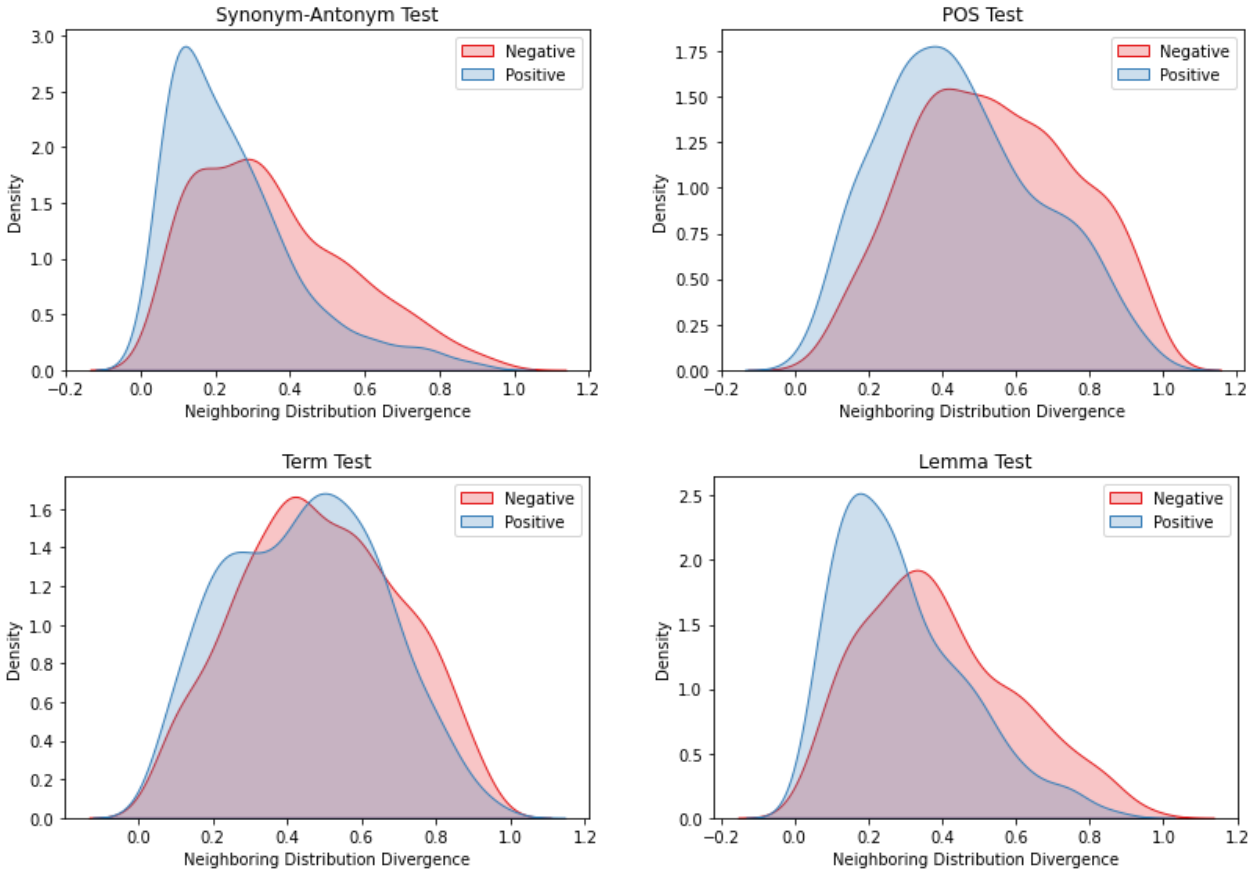}
    \caption{Distribution of bounded NDD (Hellinger distance) on semantic difference tests.}
    \label{fig:helldist}
\end{figure*}

\begin{figure*}
    \centering
    \includegraphics[width=0.99\textwidth]{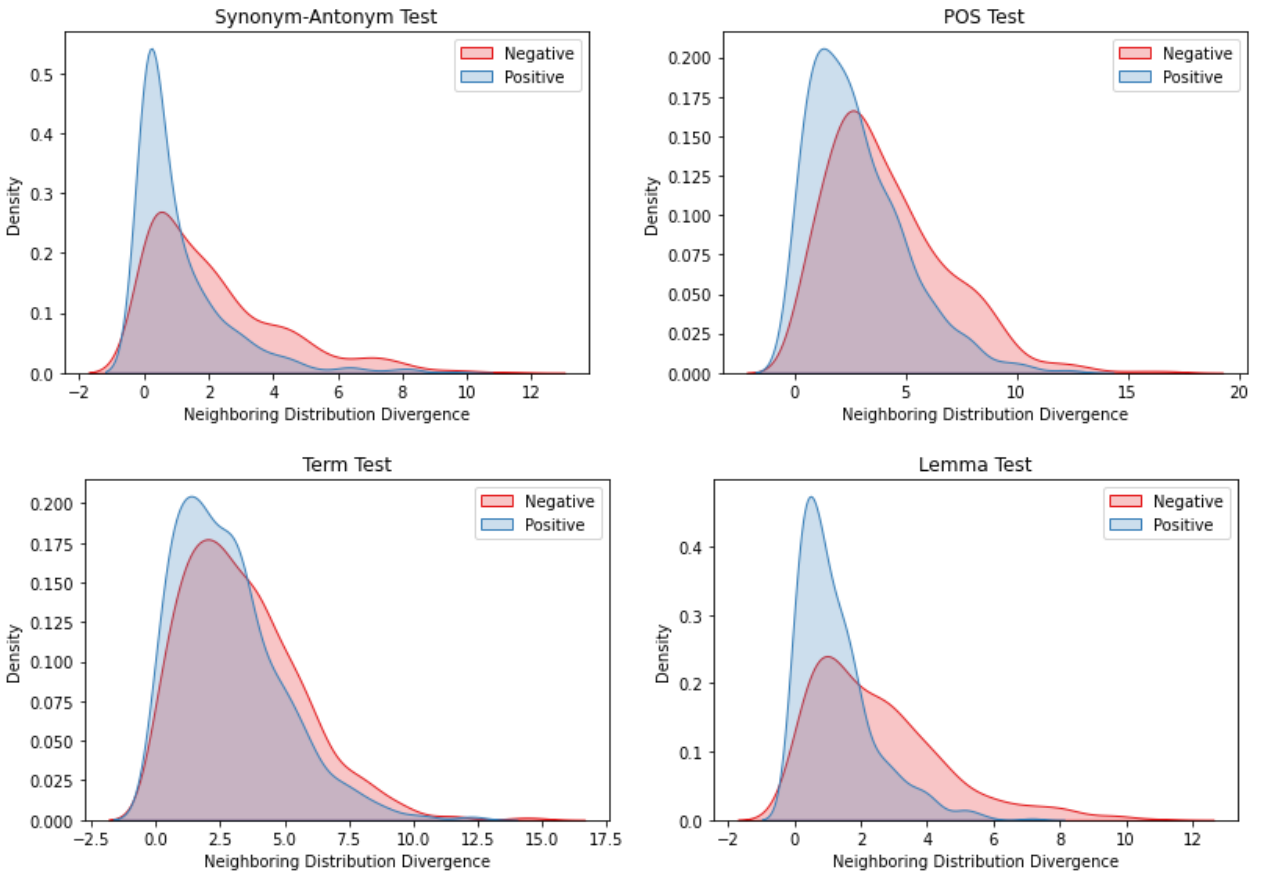}
    \caption{Distribution of unbounded NDD (KL divergence) on semantic difference tests.}
    \label{fig:kldiv}
\end{figure*}

To more specifically present how NDD is sensitive to semantic differences, we depict the distribution of bounded (Hellinger distance-based) and unbounded (KL divergence-based) NDD in Figures~\ref{fig:helldist} and~\ref{fig:kldiv}.

\clearpage

\clearpage

\section{Compression Cases}
\label{apdx:case}

\begin{table}[H]
    \centering
    \small
    \begin{tabular}{p{7.5cm}}
    \toprule
    \textbf{Init:} The speed limit on rural interstate highways in Illinois will be raised to 70 mph next year after Gov. Pat Quinn approved legislation Aug. 19, despite opposition from the Illinois Dept. of Transportation, state police and leading roadway safety organizations.\\
    \textbf{Edit:} The speed limit will be 70 mph despite opposition from organizations.\\
    \textbf{Gold:} The speed limit on highways in Illinois will be raised to 70 mph next year.\\
    \textbf{F1 Score = $51.9$ $(\downarrow 8.5)$} \quad \textbf{ROUGE = $53.8$}\\
    \midrule
    \textbf{Init:} New US ambassador to Lebanon David Hale presents credentials to Lebanese President Michel Sleiman in Baabda, Friday, Sept. 6, 2013.\\
    \textbf{Edit:} New US ambassador to Lebanon presents credentials to Lebanese President Michel Sleiman.\\
    \textbf{Gold:} New US ambassador presents credentials to Michel Sleiman.\\
    \textbf{F1 Score = $87.0$ $(\uparrow 28.7)$} \quad \textbf{ROUGE = $75.0$}\\
    \bottomrule
    \end{tabular}
    \caption{Examples for how automatic metrics reflect the performance of NDD-based compression. Improvement refers to comparison with unedited texts.}
    \label{tab:met_case}
\end{table}

\paragraph{Real Effect v.s. Automatic Metrics} As the compressed results for sentences can be various, automatic metrics might not be able to fully reflect the compressing ability of our algorithm. Also, as our compression follows a training-free procedure, the compressed results might not be in the same style as the annotated golden ones like the first instances in Table~\ref{tab:met_case}. Both our compressed and the golden result keep the main point that \textit{the speed limit will be 70 mphs}, preserving the semantics of the whole sentence. Nevertheless, the golden compression tends to keep some auxiliary information like the location \textit{on highways in  Illinois} and the time \textit{next year}. In contrast, NDD-based compression tends to remove that unimportant information and prevent semantics in other parts of the sentence from being unchanged. Thus, NDD-based compression still keeps \textit{despite opposition from organizations} towards the integrated semantics. In the second instance of Table~\ref{tab:met_case}, as the golden compression also removes location and time information from the sentence, our algorithm leads to a significant improvement since our compressing style matches with the annotated one. Considering that the automatic metrics may be biased due to the style of annotation, we present more cases in this section to show the capacity of our algorithm to keep semantics and fluency while removing unimportant and auxiliary components at the same time.

\begin{table}[H]
    \centering
    \small
    \begin{tabular}{p{7.5cm}}
    \toprule
    \textbf{Init:} A US\$5 million fish feed mill with an installed capacity of 24,000 metric tonnes has been inaugurated at Prampram, near Tema, to help boost the aquaculture sector of the country.\\
    \midrule
    \textbf{Iter1:} \textbf{A} US\$5 million \textbf{fish feed mill with} an installed \textbf{capacity} of \textbf{24,000} metric tonnes \textbf{has been inaugurated at Prampram}, near Tema, \textbf{to} help \textbf{boost the aquaculture sector} of the country\textbf{.}\\
    \midrule
    \textbf{Iter2:} \textbf{A} fish feed \textbf{mill} with capacity 24,000 \textbf{has been inaugurated} at Prampram \textbf{to boost} the \textbf{aquaculture sector.}\\
    \midrule
    \textbf{Final:} A mill has been inaugurated to boost aquaculture sector.\\
    \bottomrule
    \end{tabular}
    \caption{Cases for output in iterations of the NDD-based compression. \textbf{Bold: Kept components}}
    \label{tab:it_case}
\end{table}

\paragraph{Outputs from Compression Iterations} We present the intermediate outputs of our algorithm in Table~\ref{tab:it_case}. NDD-based text compression is shown to be capable of detecting and removing auxiliary components like locations or adjective spans in the sentence, for example. Also, the syntactic integrity and initial semantics are preserved in each iteration of our algorithm. There is an advantage over supervised methods as output in each iteration is still a plausible compression for the initial sentence. We can thus set some proper thresholds and iterate the compression until we get a fully satisfying output.

\newpage

\paragraph{Compressing Cases in Multiple Languages}

\begin{table}[H]
    \centering
    \small
    \begin{tabular}{p{7.5cm}}
    \toprule
    \begin{minipage}{0.49\textwidth}
    \begin{CJK}{UTF8}{gbsn}
     \textbf{Init:} 调价周期内，沙特下调10月售往亚洲的原油价格，我国计划释放储备原油，油价一度承压下跌。\\
     (Translation) During the price adjustment, Saudi scales down the price of crude oil sold to Asia in October, our country plans to release the reserved crude oil, oil price has once been under the dropping pressure.\\
     \textbf{Edit:} 调价周期内，沙特下调原油价格，我国释放储备原油。\\
     (Translation) During the price adjustment, Saudi scales down the price of crude oil, our country releases the reserved crude oil.
    \end{CJK}
    \end{minipage} \\
    \midrule
    \begin{minipage}{0.49\textwidth}
    \textbf{Init:} El comité de crisis, aseguró el presidente, ha tomado decisiones estratégicas que, por seguridad, no pueden ser reveladas pero que serán evidentes en las acciones que se ejecutarán en las próximas horas.\\
    (Translation) The crisis committee, the president assured, has made strategic decisions that, for security, cannot be disclosed but which will be evident in the actions that will be carried out in the next few hours.\\
    \textbf{Edit:} El comité de crisis ha tomado decisiones que no pueden ser reveladas pero serán evidentes en las acciones que se ejecutarán.\\
    (Translation) The crisis committee has made decisions that cannot be disclosed but will be evident in the actions to be carried out.
    \end{minipage} \\
    \midrule
    \begin{minipage}{0.49\textwidth}
    \begin{CJK}{UTF8}{ipxm}
     \textbf{Init:} 大型で非常に強い台風16号は、10月1日の明け方以降、非常に強い勢力で伊豆諸島にかなり近づく見込みです。\\
     (Translation) Very strong typhoon No.16 with a large scale is expected to closely approach to the Izu Islands with a very strong force after the dawn of October 1.\\
     \textbf{Edit:} 台風16号は伊豆諸島に近づく見込みです。\\
     (Translation) Typhoon No.16 is expected to approach to the Izu Islands.
    \end{CJK}
    \end{minipage} \\
    \bottomrule
    \end{tabular}
    \caption{Cases for NDD-based compression on sentences in Chinese, Spanish and Japanese.}
    \label{tab:xl_case}
\end{table}

Cases in Table~\ref{tab:xl_case} show our algorithm to be pretty well-performed on compression of other languages.

\end{document}